\documentclass[10pt,twocolumn,letterpaper]{article}

\usepackage{cvpr}              
\usepackage[table,xcdraw]{xcolor}

\usepackage{times}
\usepackage[accsupp]{axessibility}  
\usepackage[utf8]{inputenc} 
\usepackage[T1]{fontenc}    
\usepackage{url}            
\usepackage{booktabs}       
\usepackage{amsfonts}       
\usepackage{nicefrac}       
\usepackage{microtype}      
\usepackage{xcolor}         

\usepackage{graphicx}
\usepackage{amsmath,amssymb}
\usepackage{verbatim}
\usepackage{multirow}
\usepackage{marvosym}
\usepackage{makecell}
\usepackage{bm}
\usepackage{wrapfig}
\usepackage{caption}
\usepackage{twemojis}
\usepackage{utfsym}
\usepackage[pagebackref=true,breaklinks=true,letterpaper=true,colorlinks,bookmarks=false]{hyperref}

\usepackage{caption}
\usepackage{enumitem}
\setitemize{noitemsep,topsep=0pt,parsep=0pt,partopsep=0pt}

\def\paperID{10561} 
\def\confName{CVPR}
\def\confYear{2026}

\title{PAM: A Pose–Appearance–Motion Engine for Sim-to-Real HOI Video Generation}

\author{
Mingju Gao*\textsuperscript{1,2},  
Kaisen Yang*\textsuperscript{2},  
Huan-ang Gao\textsuperscript{2}, 
Bohan Li\textsuperscript{4,5},  
Ao Ding\textsuperscript{2},  \\
Wenyi Li\textsuperscript{2},  
Yangcheng Yu\textsuperscript{2}, 
Jinkun Liu\textsuperscript{2},
Shaocong Xu\textsuperscript{3},\\
Yike Niu\textsuperscript{2},
Haohan Chi\textsuperscript{2},
Hao Chen\textsuperscript{6},
Hao Tang\textsuperscript{1},
Yu Zhang\textsuperscript{\textdagger 4},
Li Yi\textsuperscript{2},
Hao Zhao\textsuperscript{\textdagger 2,3} \\
\textsuperscript{1}Peking University \quad
\textsuperscript{2}Tsinghua University \quad \textsuperscript{3}BAAI \quad\textsuperscript{4}SJTU \\ \textsuperscript{5}Eastern Institute of Technology \quad \textsuperscript{6}University of Cambridge
 \\
Project Page: \url{https://gasaiyu.github.io/PAM.github.io/}
}

\begin{document}
\maketitle

\footnotetext{\textsuperscript{*}Equal Contribution. \textsuperscript{\textdagger}Corresponding Author.}

\begin{abstract}

Hand–object interaction (HOI) reconstruction and synthesis are becoming central to embodied AI and AR/VR. Yet, despite rapid progress, existing HOI generation research remains fragmented across three disjoint tracks: (1) pose-only synthesis that predicts MANO trajectories without producing pixels; (2) single-image HOI generation that hallucinates appearance from masks or 2D cues but lacks dynamics; and (3) video generation methods that require both the entire pose sequence and the ground-truth first frame as inputs, preventing true sim-to-real deployment. Inspired by the philosophy of \cite{joo2018total}, we think that HOI generation requires a unified engine that brings together pose, appearance, and motion within one coherent framework. Thus we introduce PAM: a Pose–Appearance–Motion Engine for controllable HOI video generation. The performance of our engine is validated by: (1) On DexYCB, we obtain an FVD of 29.13 (vs. 38.83 for InterDyn), and MPJPE of 19.37 mm (vs. 30.05 mm for CosHand), while generating higher-resolution 480×720 videos compared to 256×256/256×384 baselines. (2) On OAKINK2, our full multi-condition model improves FVD from 68.76 → 46.31. (3) An ablation over input conditions on DexYCB shows that combining depth, segmentation, and keypoints consistently yields the best results. (4) For a downstream hand pose estimation task using SimpleHand, augmenting training with 3,400 synthetic videos (207k frames) allows a model trained on only 50\% of the real data plus our synthetic data to match the 100\% real baseline.

\end{abstract}    
\section{Introduction}

\begin{figure}
\centering
\includegraphics[width=0.5\textwidth]
{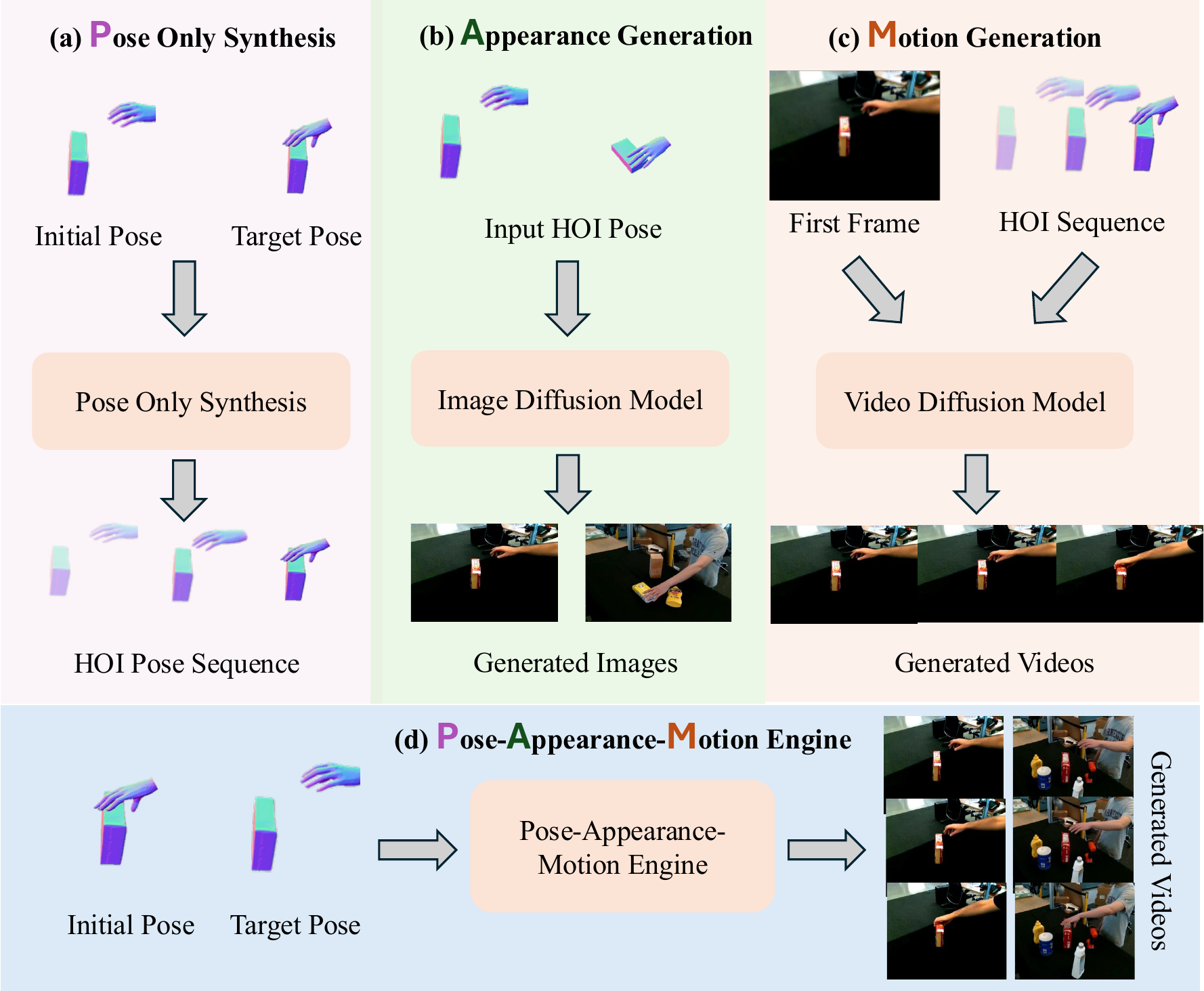}

\caption{\textbf{Overview of Four Approaches to HOI Synthesis.} \textbf{(a) Pose-Only Synthesis~\cite{zhang2024graspxl}:} This method predicts the MANO trajectories without generating pixel data; \textbf{(b) Apperance Generation~\cite{zhang2024hoidiffusion}:} This approach generates appearance based on masks or 2D cues but lacks dynamic motion; \textbf{(c) Motion Generation~\cite{pang2025manivideo, akkerman2025interdyn}:} These methods require both the full pose sequence and the ground-truth first frame as inputs, limiting their application for true sim-to-real transfer. \textbf{(d) Our Pipeline PAM:} In this approach, video generation does not rely on the first frame or the whole HOI pose sequence, allowing for the transfer of HOI pose sequences from the simulator to real-world videos.}

\vspace{-0.5cm}
\label{fig:comp}
\end{figure}

The ability to manipulate objects with our hands represents a fundamental human skill, and the computational understanding of this capability—referred to as Hand-Object Interaction (HOI) understanding—has become increasingly significant in the fields of computer vision and embodied AI. The field has seen a shift towards \textbf{data-driven} paradigms, where large-scale HOI datasets are instrumental for accurate hand pose estimation~\cite{zhou2024simple, wang2024ho, ren2025prior, yi2025estimating, yu2025dynamic, jiang2025hand, zhang2025hawor}, enabling realistic human-to-robot motion transfer~\cite{liu2025dextrack, lepert2025phantom, li2025maniptrans} and 3D modeling~\cite{liugeneralizable, yu2025dynamic, wang2025magichoi, saleem2025maskhand, zhang2025diffusion, chen2025handos, chen2025hort, potamias2025wilor, chen2022cerberus, zhao2017physics, feng2023learning, feng2024chatpose}. The critical challenge, however, lies in the data itself. Despite considerable investments in collecting real-world HOI sequences with detailed annotations~\cite{liu2022hoi4d, fu2025gigahands, yang2022oakink}, the reliance on costly and labor-intensive manual labeling poses a fundamental limitation to scalability.


The advent of deep learning and diffusion models has opened up promising avenues for scalable generation of HOI videos. However, as illustrated in Figure~\ref{fig:comp}, state-of-the-art methods~\cite{zhang2024graspxl, zhang2024hoidiffusion, pang2025manivideo, akkerman2025interdyn, sudhakar2024controlling, zhao2025analyzing} still face significant challenges: (1) Pose-only synthesis approaches are limited in generating realistic appearances, as they predict MANO trajectories without producing pixel-level details, which diminishes their practical utility. (2) Single-image HOI generation (appearance generation) methods, which rely on hallucinating appearance from masks or 2D cues, fail to capture temporal dynamics, resulting in incoherent motion when applied to real-world scenarios. (3) Video generation (motion generation) methods that require both the complete pose sequence and the ground-truth first frame are unsuitable for sim-to-real deployment, as ground-truth first frame is unavailable from the simulator. These limitations emphasize the need for innovative approaches that can generate high-quality, temporally consistent HOI videos without depending on fixed pose sequences or ground-truth first frame inputs, thus providing a more flexible and robust solution for sim-to-real transfer.

In this paper, we introduce PAM: a pioneering Pose–Appearance–Motion engine for sim-to-real HOI video generation that requires \textit{only} the initial and target poses, along with object geometry, as input. By integrating a motion-appearance diffusion process, our method bypasses the need for a conditioned first frame, thereby maximizing both motion and appearance diversity—a capability unattainable by prior work. To ensure high realism, we incorporate multiple conditions that effectively preserve fine hand details. 

Our framework consists of three core stages. The \textbf{pose sequence generation} stage produces a plausible hand motion trajectory using a pre-trained model. Next, the \textbf{appearance generation} stage synthesizes a realistic initial frame using the controllable image diffusion model Flux~\cite{flux}. This model is conditioned on a fusion of depth maps, semantic masks, and hand keypoint maps, ensuring geometric accuracy, semantic coherence, and the preservation of fine-grained hand details. The \textbf{motion generation} stage first renders the hand motion trajectory into depth, semantic, and hand keypoint sequences, and then generates videos using a controllable video diffusion model based on CogVideoX~\cite{yang2024cogvideox}. Importantly, the video model is conditioned on the same multi-modal inputs as the appearance generation stage to ensure consistency with the generated HOI pose sequence.

We evaluate our method on the DexYCB~\cite{chao2021dexycb} and OAKINK2~\cite{zhan2024oakink2} benchmarks, where it comprehensively surpasses existing approaches in video generation quality, motion plausibility, and hand pose fidelity. More importantly, as evidenced in Figure~\ref{fig:teaser}, the synthesized videos from our method provide substantial value as synthetic data. When used for training, they lead to meaningful gains in the performance of a downstream hand pose estimation model, demonstrating their effectiveness as a data augmentation tool.

We summarize our contributions as follows:
\begin{itemize}[leftmargin=*]
    \item \textbf{Minimal-Conditioning Generation:} We pioneer a pose–appearance–motion engine for sim-to-real HOI video generation that requires only sparse pose keyframes and object geometry as input, overcoming the first-frame bottleneck of prior methods.
    \item \textbf{Decoupled Generation Architecture:} We design a novel pipeline that decouples pose, appearance and motion synthesis, leveraging multi-modal conditions to achieve superior realism, controllability and diversity.
    \item \textbf{State-of-the-Art Performance and Utility:} Our method achieves superior results on established benchmarks and proves its practical value by enabling significant gains in downstream task performance through effective data augmentation.
\end{itemize}

\section{Related Works}

\subsection{Hand-Object Motion Synthesis}

Synthesizing high-fidelity hand-object motion is a fundamental challenge in computer animation and robotic grasping \citep{agarwal2023dexterous, ghosh2023imos, christen2024synh2r, li2025latenthoi, zhao2020learning}. Prevailing data-driven approaches rely on supervised learning from large-scale, well-annotated datasets \citep{jiang2021hand, karunratanakul2020grasping, dong2024hamba, pavlakos2024reconstructing, christen2022d, liu2024geneoh, li2025maniptrans, zhong2025dexgrasp, zhou2024simple, zhou2025megohand, zhang2025openhoi, huang2025hoigpt, prakash2025synthesizing}. However, the scalability of these methods is constrained by their dependence on costly and difficult-to-acquire data \citep{fan2023arctic, hampali2020honnotate, liu2022hoi4d, liu2024taco, fu2025gigahands, yang2022oakink, zhan2024oakink2, chao2021dexycb}. To circumvent this limitation, reinforcement learning (RL) has emerged as a promising alternative. Methods like \citep{christen2024synh2r, xu2023unidexgrasp} generate reference grasps before synthesizing motions, while GraspXL~\citep{zhang2024graspxl} learns a generalizable grasping policy directly in simulation, eliminating the need for predefined references. These RL-based techniques produce high-quality interaction data, forming a robust foundation for sim-to-real transfer. However, these methods lack appearance modeling.

\subsection{Controllable Video Generation}

Recent breakthroughs in video generation foundation models~\citep{yang2024cogvideox, blattmann2023stable, wan2025wan, kong2024hunyuanvideo, agarwal2025cosmos} have intensified interest in controllable generation that precisely aligns with user intent. While text-to-video and image-to-video models~\citep{agarwal2025cosmos, wan2025wan, yang2024cogvideox, singer2022make, qing2024hierarchical, guo2023animatediff, wiersma2025uncertainty, zhang2025packing, li2025uniscene} have demonstrated impressive capabilities, they often lack the granularity for specialized tasks. This has spurred research into integrating more precise control signals, such as semantic maps, depth, and camera motion. ControlNet~\citep{zhang2023adding} and its variants~\citep{gu2025diffusion, guo2024sparsectrl, li2024fairdiff, zhang2024ctrl} enable conditioning on dense inputs, while works like VideoComposer~\citep{wang2023videocomposer} fuse multiple conditions for enhanced control. Camera motion has been explicitly modeled by embedding parameters into diffusion models~\citep{he2024cameractrl, bai2025recammaster}. However, generating videos of hand-object interactions (HOI) presents a unique challenge due to the high degrees of freedom in hand motion. This demands even more enriched and specialized control mechanisms—combining semantic, geometric, and precise pose cues—to achieve the necessary fidelity and accuracy.

\subsection{Hand-Object Interaction Image \& Video Generation}

Generating Hand-Object Interaction (HOI) content is vital for understanding human activities. Prior work on \textit{HOI image generation}~\citep{hu2022hand, kwon2024graspdiffusion, pelykh2024giving, wang2025realishuman, ye2023affordance, zhang2024hoidiffusion, chen2025foundhand} typically conditions on 2D signals like segmentation masks and keypoints. However, these static methods cannot capture the dynamic nature of interactions. Recently, several studies~\citep{sudhakar2024controlling, pang2025manivideo, akkerman2025interdyn, dang2025svimo, ye2025textsc, fan2025re, zhao2025taste} have explored \textit{HOI video generation}. InterDyn~\citep{akkerman2025interdyn} conditions on hand mask sequences via ControlNet~\citep{zhang2023adding}, but under-utilizes the rich conditions available from simulators. ManiVideo~\citep{pang2025manivideo} introduces an occlusion-aware representation but requires human appearance data, which is not available from simulators like GraspXL~\citep{zhang2024graspxl}. More critically, these methods primarily focus on generation quality and have not thoroughly investigated the \textit{downstream utility} of their synthesized data, which is essential for validating practical impact beyond perceptual metrics.
\section{The Proposed Method}

\begin{figure*}
\centering
\includegraphics[width=0.8\textwidth]{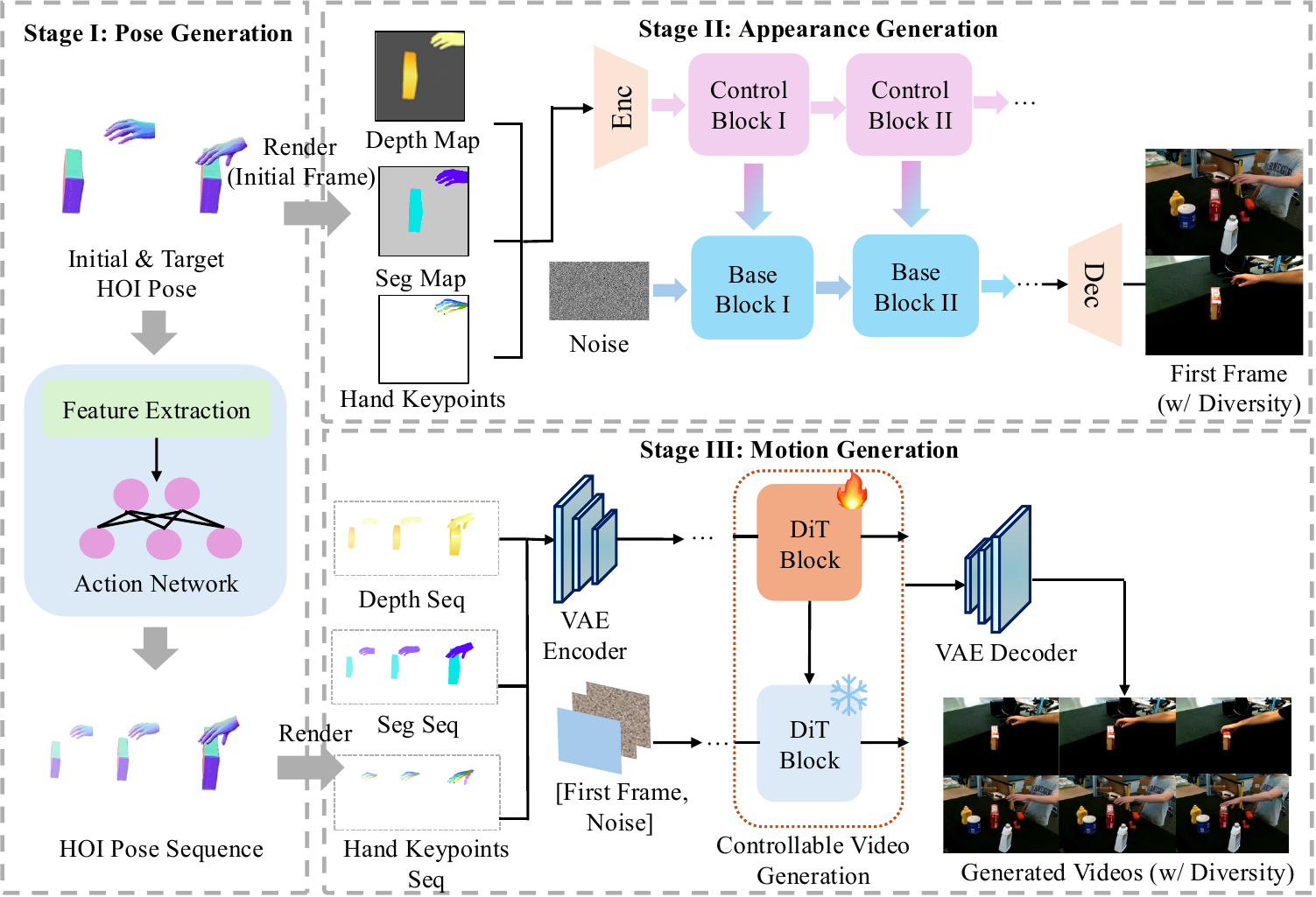}
\caption{\textbf{Overview of our three-stage generation pipeline.} (1) \textbf{Pose Generation:} A pretrained pose generation model generates the intermediate hand-object interaction (HOI) poses based on the initial and target poses, along with the object mesh. (2) \textbf{Appearance Generation:} A controllable image diffusion model synthesizes the first frame of the video, conditioned on multi-modal inputs (depth maps, semantic masks, and keypoint annotations). (3) \textbf{Motion Generation:} The generated HOI sequence and the first frame are rendered into a full video sequence by a video diffusion model, conditioned on the same multi-modal inputs used in the appearance generation stage.}

\vspace{-0.4cm}
\label{fig:main}
\end{figure*}
\subsection{Overview}

Figure~\ref{fig:main} illustrates our method.
Conditioned on an initial MANO~\citep{romero2022embodied} hand pose $\mathbf{h}_0\!\in\!\mathbb{R}^{51\times3}$, an object mesh $\mathbf{m}$ without appearance, an initial 6-DoF object pose $\mathbf{o}_0\!\in\!\mathbb{R}^{6}$, and a target hand pose $\mathbf{h}_T\!\in\!\mathbb{R}^{51\times3}$, our generative model
\begin{equation}
    f_{\boldsymbol{\theta}}: (\mathbf{h}_0,\mathbf{m}, \mathbf{o}_0,\mathbf{h}_T)\;\rightarrow\;\{I_t\}_{t=0}^{T}
\end{equation}
produces a photo-realistic video that (i) begins with $\mathbf{h}_0$, (ii) ends with $\mathbf{h}_T$, and (iii) depicts a temporally-coherent grasp-to-place motion. All hand poses are parameterised by global translation + rotation plus joint angles; frames $I_t$ are RGB images.

Jointly modeling pose, appearance and motion is inherently challenging due to the high-dimensional spatio-temporal manifold~\citep{guo2024i4vgen}. To address this, we decompose the generation process into three distinct stages:

\begin{enumerate}
    \item \textbf{Pose Generation} --- A pretrained pose generation model synthesizes aligned hand-object pose sequences, $\{\mathbf{h}_t,\mathbf{o}_t\}_{t=0}^{T}$, from the initial and target hand-object poses and object mesh, $(\mathbf{h}_0,\mathbf{o}_0, \mathbf{m})$ (Sec.~\ref{Sec:first}).
    \item \textbf{Appearance Generation} --- A controllable image diffusion model generates the first frame, $I_0$, conditioned on the initial HOI pose and object mesh (Sec.~\ref{Sec:second}).
    \item \textbf{Motion Generation} --- The synthetic HOI pose sequences and the first frame are injected into a video diffusion model, which animates $I_0$ into a photo-realistic video clip $\{I_t\}_{t=0}^T$ (Sec.~\ref{Sec:third}).
\end{enumerate}

\subsection{Stage I: Pose Generation Stage}
\label{Sec:first}

In the first stage, we aim to generate an interpolated hand-object interaction (HOI) pose sequence, transitioning from the initial to the target pose while incorporating the object mesh. As shown in Figure~\ref{fig:main}, we employ GraspXL~\citep{zhang2024graspxl}, a pretrained model for hand-object interaction tasks. GraspXL takes as input the initial and target MANO hand pose $\mathbf{h}_0$, the 6-DoF object pose $\mathbf{o}_0$, and the object mesh $\mathbf{m}$, and produces temporally coherent trajectories $\{\mathbf{h}_t, \mathbf{o}t\}_{t=0}^{T}$ for both hand and object poses. This ensures smooth interpolation while maintaining the physical consistency of the hand-object interaction, providing the foundation for subsequent video generation stages.

\subsection{Stage II: Appearance Generation Stage}
\label{Sec:second}

\textbf{Bridge Conditions for Sim-to-Real HOI Video Synthesis:} The primary objective of this work is to enhance the visual quality of simulated videos while preserving other conditions, thereby bridging the gap between simulation and real-world scenarios. By incorporating both geometric information (e.g., depth maps) and semantic data (e.g., segmentation masks) from the simulator, we seek to accurately reconstruct the visual representation of scenes and objects, while ensuring consistency across all other conditions. However, relying solely on these two data types proves insufficient for accurately generating Hand-Object Interactions (HOI) images or videos. This limitation stems from the complexity and high degree of freedom inherent in hand movements, which cannot be fully captured by geometric and semantic data alone. Specifically, these conditions fail to account for critical details, such as the number of fingers and their individual poses. To address this challenge, we introduce an additional condition—hand keypoint sequences, as proposed by \citep{zhang2024hoidiffusion}—to enable more precise and accurate hand pose generation. This approach facilitates the generation of realistic hand poses, thereby enhancing the overall realism of the interaction. In section~\ref{sec:exp_main} and~\ref{sec:exp_abl}, we explore the influence of every condition.

We fine-tune Flux~\citep{flux} with a ControlNet~\citep{zhang2023adding} fork that accepts depth $D_0$, segmentation $S_0$ and hand-keypoint image $K_0$ ($H\!\times\!W\!\times\!3$ each).
All cues are VAE-encoded to $\frac{H}{8}\!\times\!\frac{W}{8}\!\times\!16$ latents, concatenated channel-wise and Injected into two layers of DiT~\citep{peebles2023scalable} blocks, with weights initialized from the first two layers of original Flux.:
\begin{equation}
    f_l = f_l + \mathcal{Z}(f'_l),
    \label{eq:control}
\end{equation}
where $f_l$ is the output of the $l$-th layer of the original DiT~\citep{peebles2023scalable} blocks, and $f'_l$ is the output of the $l$-th layer of the duplicated DiT blocks whose input is the concatenated conditions. Here, $l \in \{0, 1\}$, and $\mathcal{Z}$ represent the zero-convolution layer, which is a $1 \times 1$ convolution with all parameters initialized to zero. During training, only the parameters of ControlNet are updated. 

\subsection{Stage III: Motion Generation Stage}
\label{Sec:third}

To generate the target video sequence, we combine the generated HOI pose sequence with a controllable video diffusion model. As shown in Figure~\ref{fig:main}, depth maps, instance-level segmentation masks, and 2-D hand keypoint images are rasterized at each frame from Stage I. These conditions are then encoded into a latent tensor of shape $\mathbb{R}^{\frac{T+1}{4}\times\frac{H}{8}\times\frac{W}{8}\times 16}$ using a pretrained video VAE. Next, the latent representations are concentrated along channel dimensions and injected into CogVideo-X through 12 duplicate DiT blocks, as described in Eq.~\ref{eq:control}. To prevent over-reliance on any single modality during training, each cue is randomly masked with a probability of 0.2.
\section{Experiment}

\subsection{Experiment Settings}
\label{sec:exp_setting}

\renewcommand{\arraystretch}{1.1}  
\begin{table*}[tbp]
    \centering
    \resizebox{1.75\columnwidth}{!}{
        \begin{tabular}{ccccccccc}
            \toprule
            \textbf{Method} & \textbf{Venue} & \textbf{FVD (↓)} & \textbf{MF (↑)} & \textbf{LPIPS (↓)} & \textbf{SSIM (↑)} & \textbf{PSNR (↑)} & \textbf{MPJPE (↓)} & \textbf{Resolution}  \\
            \midrule
            CosHand~\cite{sudhakar2024controlling} & ECCV'24 &  58.51 &  0.591 & 0.139 & 0.767 & 23.20 & 30.05 & 256 x 256 \\
            InterDyn~\cite{akkerman2025interdyn} & CVPR'25 &38.83 &  0.680 & 0.119 & 0.848 & 24.86 & - & 256 x 384 \\ 
            ManiVideo~\cite{pang2025manivideo} & CVPR'25 & - & - & 0.079 & \underline{0.913} & \underline{30.10} & 57.30 & - \\
            \hline
            Ours w/ all &  - &\textbf{29.13} &  \textbf{0.712} & \textbf{0.069} & \textbf{0.914} & \textbf{30.17} & \textbf{19.37}  & 480 x 720\\
            \bottomrule
        \end{tabular} }

    \caption{\textbf{Quantitative comparison on DexYCB dataset.} Our method is evaluated against CosHand, InterDyn, and ManiVideo. Results for InterDyn and ManiVideo are taken from their original papers. For fair comparison, CosHand was fine-tuned on the s0-split training set identical to ours. Our approach achieves state-of-the-art performance across all metrics (FVD, LPIPS, MF, MPJPE) while generating high-resolution 480x720 videos.}
    \label{tab:main_dexycb}
\end{table*}

\label{sec:exp_main}

\begin{table*}[tbp]
    \centering
    \resizebox{1.4\columnwidth}{!}{%
        \begin{tabular}{cccccccc}
            \toprule
            \textbf{Method} & \textbf{FVD (↓)} &  \textbf{MF (↑)} & \textbf{LPIPS (↓)} & \textbf{SSIM (↑)} & \textbf{PSNR (↑)} & \textbf{MPJPE (↓)}  \\
            \midrule
            CosHand~\cite{sudhakar2024controlling} & 68.76 &  0.651 & 0.156 & 0.765 & 23.84 & 14.49 \\
            Ours w/ seg & \underline{48.97} & \underline{0.708} & \underline{0.084} & 0.831 & 25.76 & 9.61  \\
            Ours w/ depth & 50.85 & 0.702 & 0.086 & \underline{0.845} & \underline{26.98} & 10.07    \\
            Ours w/ hand & 52.41 &  0.671 & 0.113 & 0.838 & 25.66 & \underline{8.01} \\
            Ours w/ all & \textbf{46.31} &  \textbf{0.777} & \textbf{0.081} & \textbf{0.851} & \textbf{28.36} & \textbf{7.01} \\
            \bottomrule
        \end{tabular}
    }

    \caption{\textbf{Quantitative results on the OAKINK2 dataset.} Comparison of our method with CosHand. For a fair evaluation, both models are trained on the same dataset. Our approach achieves state-of-the-art performance, outperforming CosHand across all evaluated metrics.}

    \label{tab:main_oakink2}
\end{table*}

\textbf{Datasets and Data Processing.} We evaluate our method on two standard benchmarks for HOI video generation: DexYCB~\citep{chao2021dexycb} and OAKINK2~\citep{zhan2024oakink2}. For DexYCB, we adopt the s0-split, comprising 6,400 training and 1,600 validation videos. Due to the scale of OAKINK2, we use a curated subset of 8,000 video clips (each 49 frames long), split into 6,400 for training and 1,600 for validation. The conditions for our model—depth maps, semantic masks, and hand keypoints—are derived as follows: depth maps are estimated using DepthCrafter~\citep{hu2025depthcrafter}, while semantic and keypoint information are obtained directly from the dataset annotations.

\begin{figure*}
\centering
\includegraphics[width=0.95\textwidth]{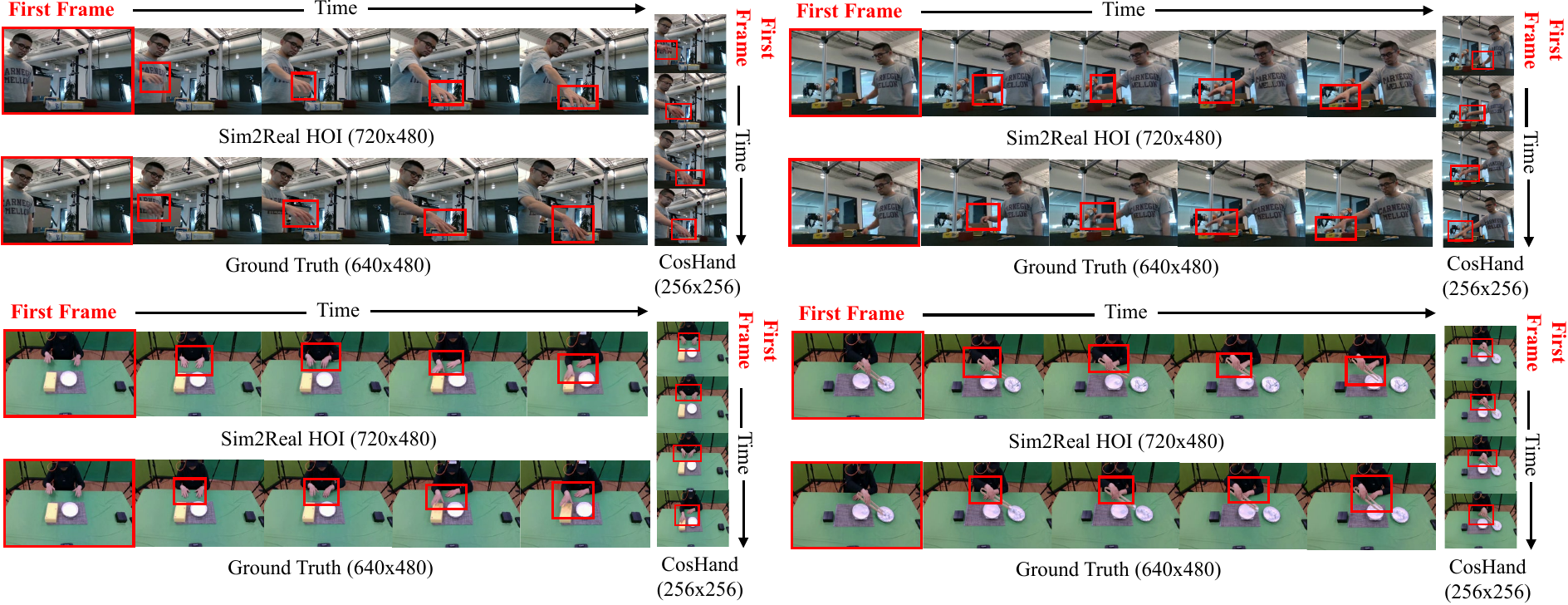}
\caption{\textbf{Qualitative comparison against CosHand.} Example results on DexYCB and OAKINK2 highlight the strengths of our method in two key areas: (1) higher visual fidelity in both foreground and background generation, and (2) improved geometric accuracy of the synthesized hand poses.}

\vspace{-0.3cm}
\label{fig:quati}
\end{figure*}

\textbf{Evaluation Metrics.} We employ a comprehensive set of metrics to evaluate our method from four perspectives:
\begin{itemize}[leftmargin=*]
    \item \textbf{Image Quality:} We assess perceptual quality using Structural Similarity Index Measure (SSIM), Learned Perceptual Image Patch Similarity (LPIPS) \citep{zhang2018unreasonable} and Peak Signal-to-Noise Ratio (PSNR).
    \item \textbf{Spatio-temporal Coherence:} We adopt Fréchet Video Distance (FVD) \citep{Unterthiner2018TowardsAG} to evaluate overall video realism, using the implementation from \citep{skorokhodov2022stylegan}.
    \item \textbf{Motion Fidelity:} We use the Motion Fidelity (MF) metric \citep{yatim2024space} to quantify dynamic accuracy. For each video, we sample 100 foreground points (on hands/objects), track them using CoTracker3 \citep{karaev2024cotracker3}, and compare the trajectories between generated and ground-truth videos. For a ground-truth tracklet $\mathcal{T} = \{\mathbf{\tau}_1, \dots, \mathbf{\tau}_T\}$ and a generated tracklet $\mathcal{\tilde{T}} = \{\mathbf{\tilde{\tau}}_1, \dots, \mathbf{\tilde{\tau}}_T\}$ where $\mathbf{\tau}_t \in \mathbb{R}^2$, MF is defined as:
    \begin{equation}
        \text{MF} = \frac{1}{|\mathcal{\tilde{T}}|} \sum_{\tilde{\tau} \in \mathcal{\tilde{T}}} \max_{\tau \in \mathcal{T}} \textbf{corr}(\tau, \tilde{\tau}) + \frac{1}{|\mathcal{T}|} \sum_{\tau \in \mathcal{T}} \max_{\tilde{\tau} \in \mathcal{\tilde{T}}} \textbf{corr}(\tau, \tilde{\tau}).
    \end{equation}
    The correlation between two tracks is computed as:
    \begin{equation}
        \textbf{corr}(\tau, \tilde{\tau}) = \frac{1}{F} \sum_{k=1}^F \frac{\mathbf{v}_k \cdot \mathbf{\tilde{v}}_k}{\|\mathbf{v}_k\| \|\mathbf{\tilde{v}}_k\|},
    \end{equation}
    where $\mathbf{v}_k = (v_k^x, v_k^y)$ and $\mathbf{\tilde{v}}_k = (\tilde{v}_k^x, \tilde{v}_k^y)$ are the displacement vectors at the $k$-th frame for tracks $\tau$ and $\tilde{\tau}$, respectively.
    \item \textbf{Hand Pose Accuracy:} We report Mean Per-Joint Position Error (MPJPE) in millimeters \citep{fan2023arctic}, measuring the average Euclidean distance between the 21 predicted and ground-truth hand joints after root alignment. Lower MPJPE indicates better pose estimation accuracy. We utilize Hamer~\cite{pavlakos2024reconstructing} to estimate the hand joints in the generated videos
\end{itemize}

\subsection{Main Results}


\textbf{Baselines.} We compare our method against state-of-the-art HOI video generation approaches: ManiVideo~\citep{pang2025manivideo}, InterDyn~\citep{akkerman2025interdyn}, and CosHand~\citep{sudhakar2024controlling} on the DexYCB dataset~\citep{chao2021dexycb}. For ManiVideo and InterDyn, we report results directly from their original publications (omitting metrics for which results were unavailable due to these methods not being open-source). For CosHand, we use the official implementation and fine-tune it on the DexYCB s0-split training set for a fair comparison. We also evaluate on OAKINK2, comparing against a similarly fine-tuned CosHand model. All baselines are evaluated in an image-to-video setting where the ground-truth first frame is provided, as this is required by these methods.


\textbf{Quantitative Comparisons.}
Our quantitative evaluation (Tables~\ref{tab:main_dexycb}, ~\ref{tab:main_oakink2}) demonstrates that our method achieves state-of-the-art performance across most metrics. We attribute this superiority to our multi-conditioning strategy, which provides the diffusion model with rich geometric and semantic cues (depth, masks, keypoints) to jointly optimize for visual realism and pose accuracy. In contrast, baseline methods exhibit limitations: InterDyn, ManiVideo, and CosHand rely on more limited conditioning signals or are built upon foundation models that struggle to capture the intricacies of hand-object interactions, leading to suboptimal performance.

\begin{table*}[tbp!]
    \centering
    \resizebox{1.4\columnwidth}{!}{%
        \begin{tabular}{cccccccc}
            \toprule
            \textbf{Conditions} & \textbf{FVD (↓)} & \textbf{MF (↑)} & \textbf{LPIPS (↓)} & \textbf{SSIM (↑)} & \textbf{PSNR (↑)} & \textbf{MPJPE (↓)}  \\
            \midrule
            Seg & 33.23 &  0.695 & 0.077 & 0.900 & 29.27 & 21.14 &  \\
            Depth & 30.00 &  0.703 & \underline{0.070} & \underline{0.906} & 29.15 & 23.16   & \\
            Hand & 33.41 &  \textbf{0.713} & 0.086 & 0.901 & 29.07 & 20.70  & \\
            \hline
           Depth, Hand & 29.62 & 0.711 & 0.071 & 0.899 & 29.95 & 20.46    \\
            Seg, Hand & 29.53 & 0.711 & 0.073 & 0.902 & 29.57 & \underline{19.92}    \\
            Depth, Seg & \underline{29.32} &  \underline{0.712} & 0.071 & \underline{0.906} & \textbf{30.60} & 22.51 \\
            \hline
            Depth, Hand, Seg & \textbf{29.13} & \underline{0.712} & \textbf{0.069} & \textbf{0.914} & \underline{30.17} & \textbf{19.37} \\
            \bottomrule
        \end{tabular}
    }
    \caption{Ablation study on input conditions on DexYCB dataset.}
\vspace{-0.5cm}
    \label{tab:abl}
\end{table*}

\textbf{Qualitative Comparisons.}
As shown in Figure~\ref{fig:quati}, our method generates visually superior results compared to CosHand, even when CosHand is fine-tuned on the same training data. We identify two primary limitations in CosHand: (1) its reliance on hand masks as the sole conditioning signal provides insufficient geometric guidance for reconstructing precise hand poses, and (2) its lack of explicit temporal modeling mechanisms leads to inconsistent frame-to-frame outputs. In contrast, our approach addresses these issues by leveraging a video diffusion foundation model equipped with temporal attention to enforce coherence across frames. Furthermore, the use of hand keypoint maps as a conditioning signal explicitly preserves the structural details of hand configurations, resulting in more accurate and smooth video sequences.

\label{sec:exp_abl}

\begin{figure*}
\centering
\includegraphics[width=0.8\textwidth]{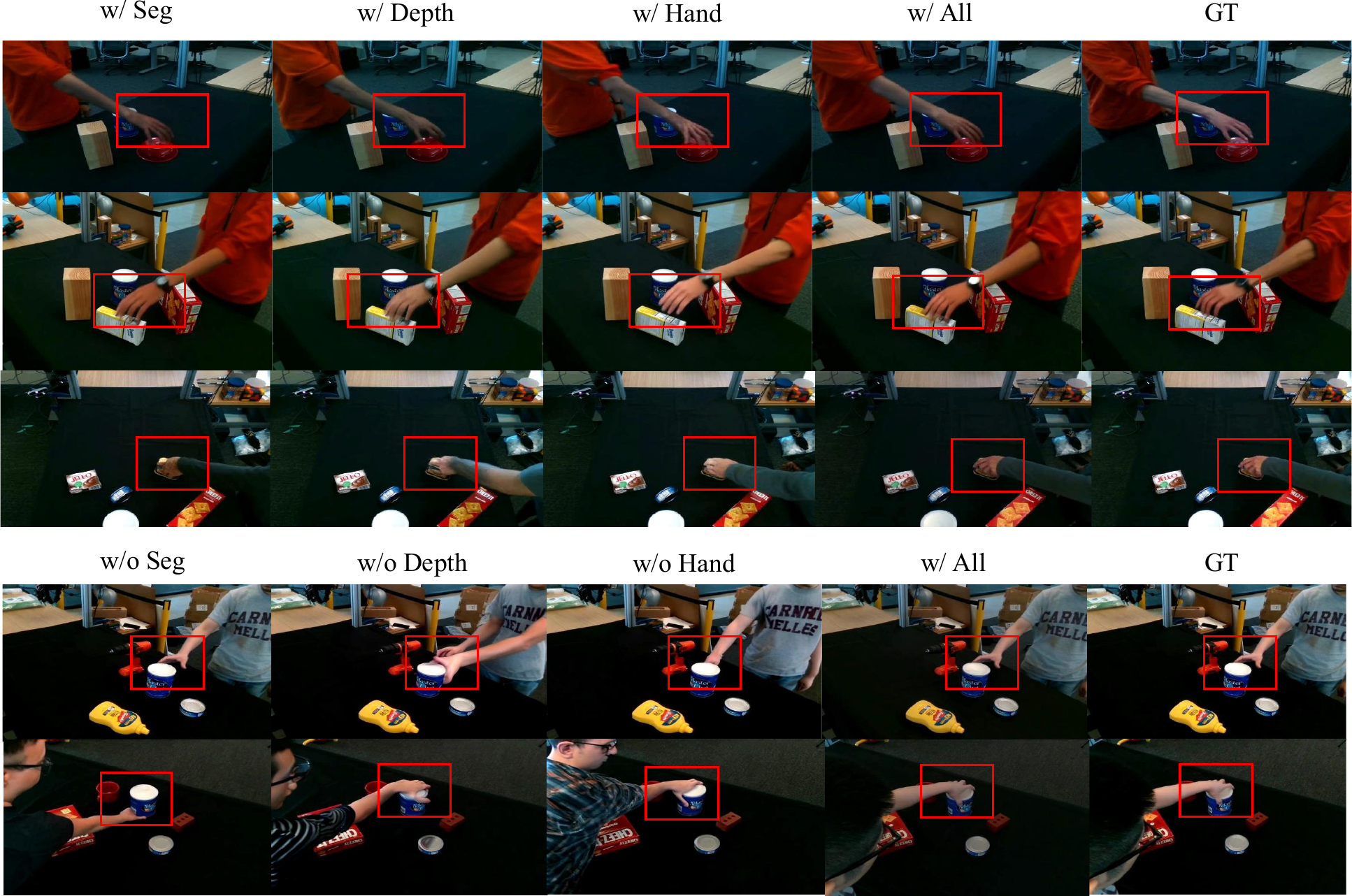}
\caption{\textbf{Ablation study on input conditions on DexYCB dataset.} 
}
\vspace{-0.5cm}
\label{fig:abl}
\end{figure*}

\subsection{Ablation Studies on Input Conditions}

We conduct an ablation study on the DexYCB dataset to evaluate the contribution of different input conditions. The results (Tables \ref{tab:main_dexycb}, \ref{tab:main_oakink2}, and \ref{tab:abl}) yield three key observations: 
\begin{itemize}[leftmargin=*]
    \item The performance improves with an increasing number of conditions, validating the effectiveness of our multi-condition design.
    \item Even when using the same segmentation mask condition as CosHand and InterDyn, our method achieves superior results, demonstrating the advantage of our pipeline.
    \item While using only hand keypoints yields low MPJPE (due to explicit pose supervision), it underperforms on other metrics due to the lack of broader geometric and semantic context. This highlights the necessity of combining detailed local cues (keypoints) with global scene understanding (depth, semantics) for optimal performance.
\end{itemize}
Visual results in Figure~\ref{fig:abl} further support these findings: using all conditions produces accurate poses; semantic masks or depth maps alone lead to pose inaccuracies; and keypoints alone degrade appearance quality.

\subsection{Diverse Sim-to-Real Transfer}
\begin{figure*}
\centering
\includegraphics[width=0.8\textwidth]{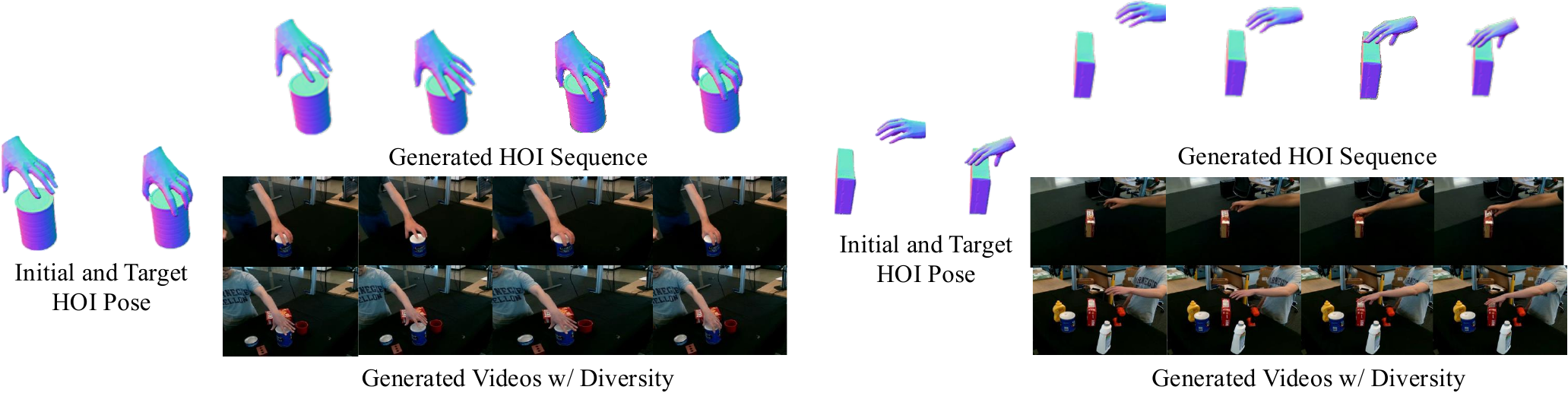}
\caption{\textbf{Sim-to-real transfer results.} Our pipeline can generate realistic videos given initial and target states with diversity. 
}
\vspace{-0.2cm}
\label{fig:sim}
\end{figure*}

\begin{table*}[htb]
    \centering
    \resizebox{1.4\columnwidth}{!}{%
        \begin{tabular}{ccccc}
            \toprule
            \textbf{Setting} & \textbf{PA-MPJPE (↓)} & \textbf{PA-MPVPE (↓)} & \textbf{F-Score@05 (↑)} & \textbf{F-Score@15(↑)} \\
            \midrule
             All real data & 5.5 & \underline{5.5} & 0.7953 & \underline{0.9899} \\ 
            All gen. data & 8.2 & 8.1 & 0.6274 & 0.9626 \\ 
            All gen. + 25\% real data & 6.1 & 6.0 & 0.7512 & 0.9851 \\ 
             All gen. + 50\% real data & 5.5 & \underline{5.5} & 0.8001 & 0.9879  \\ 
             All gen. + 75\% real data &  \underline{5.4} & \textbf{5.3} & \underline{0.7984} &  \underline{0.9899} \\ 
             All gen. + 100\% real data & \textbf{5.3} & \textbf{5.3} & \textbf{0.8025} & \textbf{0.9904} \\ 
            \bottomrule
        \end{tabular}
    }
    \caption{Downstream task evaluation on SimpleHand~\citep{zhou2024simple}.}
    \label{tab:down}
\end{table*}

\begin{figure*}[!htb]
\centering
\includegraphics[width=0.8\textwidth]{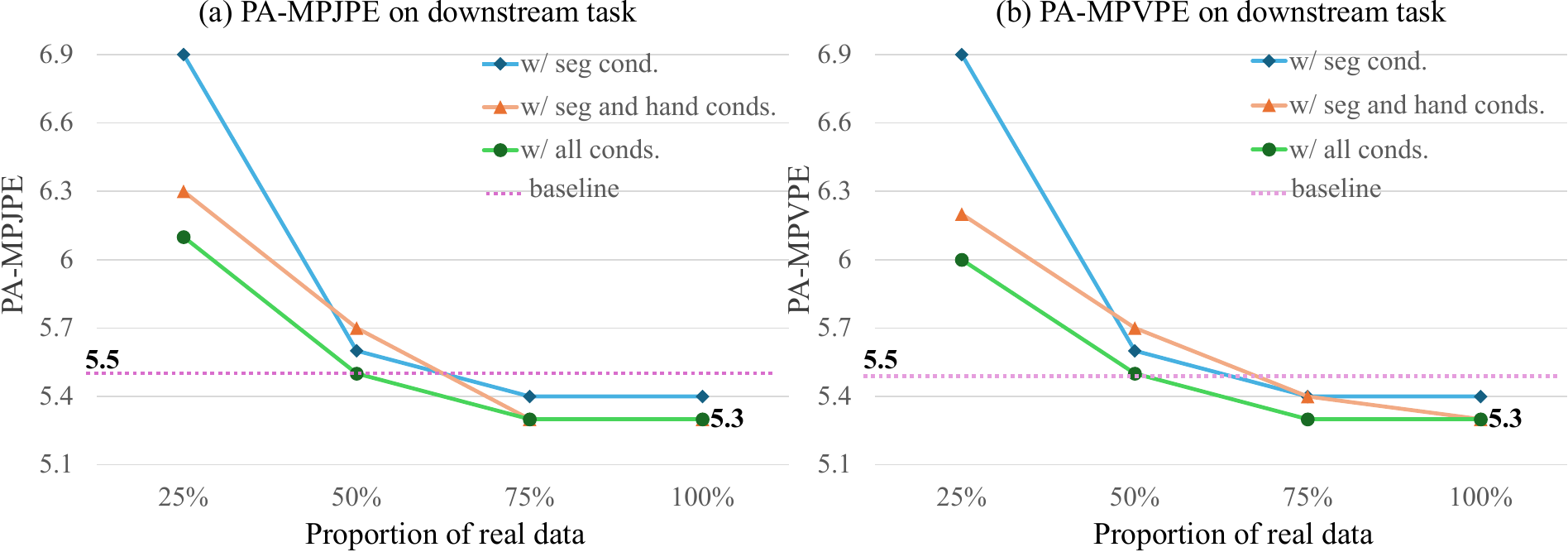}

\caption{Data augmentation analysis with varying ratios of real data. We augment different portions of the DexYCB training set (25\%, 50\%, 75\%, 100\%) with our generated synthetic data. The \textbf{baseline} (dashed line) indicates performance when training solely on 100\% of the real DexYCB data without synthetic augmentation.}

\label{fig:down}
\vspace{-0.3cm}
\end{figure*}

We conduct a sim-to-real transfer experiment to validate the effectiveness of our full pipeline. For this, we use GraspXL~\citep{zhang2024graspxl} as the hand motion generator, owing to its superior performance and strong generalization capabilities. Using objects from the DexYCB dataset, we randomly initialize both the hand and object poses, along with a target hand pose. GraspXL then generates the intermediate motion sequence, which is used to render the necessary conditions—depth, semantic masks, and keypoints—for our video generation model. As demonstrated in Figures~\ref{fig:teaser} and~\ref{fig:sim}, our method successfully synthesizes \textbf{diverse, realistic videos} with varying subjects and backgrounds, using minimal input that includes only the initial and target poses, along with the object geometry. This capability arises from our decoupled generation architecture, which effectively combines the motion prior from GraspXL with the appearance modeling of our diffusion model. The utility of these synthesized videos for downstream tasks is explored in Section~\ref{sec:down}.

\subsection{Downstream Task Validation}
\label{sec:down}

To evaluate the utility of our generated videos, we employ them for data augmentation in a hand pose estimation task. We use SimpleHand~\citep{zhou2024simple} as the pose estimation model, which regresses MANO parameters~\citep{romero2022embodied} from a single image. Our pipeline, trained on DexYCB, generates 3,400 video sequences (207,400 frames) for augmentation. We combine this synthetic data with varying subsets (25\%, 50\%, 75\%, 100\%) of the original DexYCB s0-split training set (406,888 frames). All models are evaluated on the DexYCB validation set using four metrics: Procrustes-Aligned Mean Per-Joint Position Error (PA-MPJPE), Procrustes-Aligned Mean Per-Vertex Position Error (PA-MPVPE), and F-Score. PA-MPJPE/PA-MPVPE measure the average Euclidean distance (in mm) after Procrustes alignment between the predicted and ground-truth joints/vertices, respectively. 
The PA-MPJPE metric used here differs from the one in Table~\ref{tab:main_dexycb}. The MPJPE in this context measures data efficiency for downstream tasks, whereas the MPJPE in Table~\ref{tab:main_dexycb} refers to the hand accuracy of the generated videos.

The quantitative results (Table~\ref{tab:down}) demonstrate that incorporating our generated data consistently improves hand pose estimation accuracy across all metrics. Figure~\ref{fig:down} reveals two key trends: (1) model performance improves monotonically with the amount of real data, and (2) most notably, using only 50\% of the real data augmented with our synthetic samples achieves competitive performance with the 100\% real data baseline. This indicates that our synthetic data can effectively compensate for reduced real data volume. Furthermore, the superior performance achieved using videos generated with multiple conditions validates the importance of our multi-conditioning approach for producing diverse and useful training data.

\begin{figure}
\centering
\includegraphics[width=0.5\textwidth]{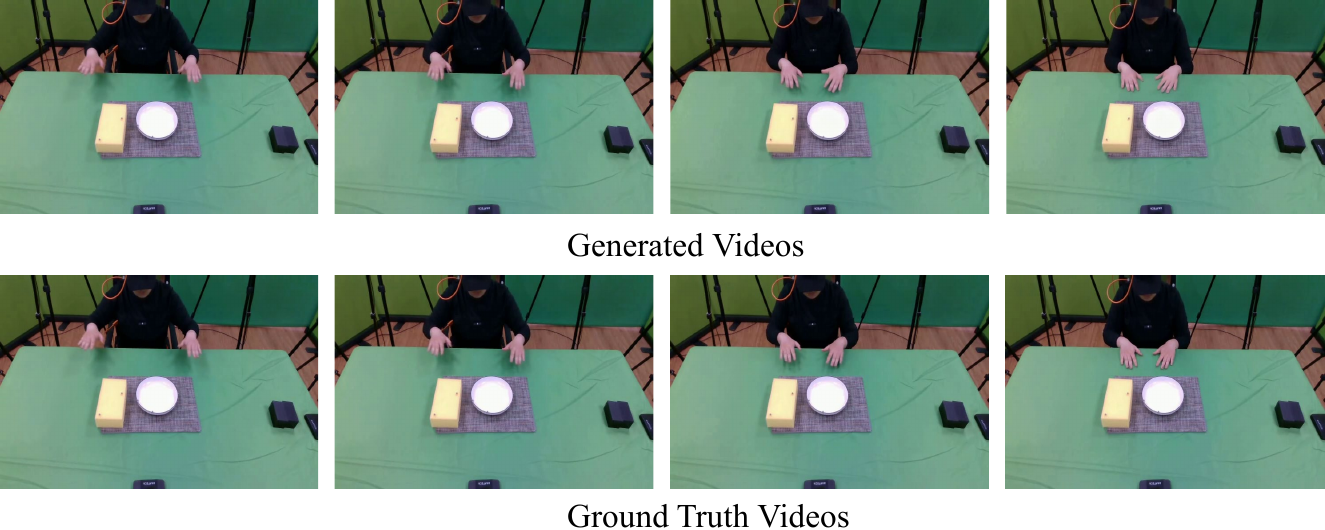}
\caption{Zero-shot result on OAKINK2 dataset for i2v task. We use the weight trained on DexYCB dataset.}
\label{fig:zero}
\vspace{-0.7cm}
\end{figure}

\subsection{Zero-Shot Results}

To evaluate the generalizability of our approach, we test our video diffusion model trained on the DexYCB dataset (single-hand interactions) directly on the OAKINK2 dataset (bimanual interactions) in a zero-shot image-to-video setting. As shown in Figure~\ref{fig:zero}, our method generates plausible videos that maintain reasonable alignment with ground-truth hand poses and visual details, despite the significant domain shift. This cross-dataset generalization capability can be attributed to our use of pretrained video diffusion model weights as a strong foundation, combined with the ControlNet mechanism~\citep{zhang2023adding}, which helps preserve the model's original generation quality while adapting to new conditioning signals.

\section{Conclusion}


This paper proposed a framework that addresses the challenge of generating realistic HOI videos from minimal pose inputs. Our decoupled, multi-condition architecture produces superior results in both perceptual quality and geometric accuracy, and demonstrates practical utility through enhanced downstream task performance. While our method shows strong generalization, future work could explore extending it to more complex object interactions or unifying the motion and appearance stages into an end-to-end model. We believe our contributions provide a solid foundation for future research in generative models for embodied AI.

\section*{Acknowledgments} This study is partially supported by Beijing Natural Science Foundation, QY25047.

{
    \small
    \bibliographystyle{ieeenat_fullname}
    \bibliography{main}
}

\newpage

\section{Implementation Details}

\begin{table}[h]
\centering
\begin{tabular}{c|cccc}
\toprule
\textbf{Resource Usage} & \textbf{Stage-I} & \textbf{Stage-II} & \textbf{Stage-III} \\
\midrule
Memory (GB)             & 0.03             & 41.4              & 30.3              \\
Time (s)                & 19.3             & 36.1              & 245.7             \\
\bottomrule
\end{tabular}
\caption{Resource Usage across Different Stages}
\label{sup:resources}
\end{table}

\begin{table}[t]
\centering
\caption{Ablation Study on Masking Probability and Input Condition Quality}
\label{tab:mask}
\resizebox{\linewidth}{!}{
\begin{tabular}{lccc} 
\hline
\textbf{Settings}                      & \textbf{FVD ($\downarrow$)} & \textbf{PSNR ($\uparrow$)} & \textbf{MPJPE ($\downarrow$)} \\ \hline
0 Mask Prob + Clean Cond               & 28.56                       & 30.99                      & 19.01                         \\ 
0 Mask Prob + Noisy Cond               & 34.58                       & 27.11                      & 23.67                         \\ 
0.2 Mask Prob + Clean Cond (\textbf{Ours}) & \textbf{29.13}          & \textbf{30.17}             & \textbf{19.37}                \\ 
0.2 Mask Prob + Noisy Cond             & 30.45                       & 29.67                      & 20.31                         \\ \hline
\end{tabular}
} 
\end{table}

\begin{table}[hbp]
\centering
\caption{Ablation on Stage-I Method}
\begin{tabular}{c|ccc}
\toprule
Method (Stage-I) & FVD ($\downarrow$) & MF ($\uparrow$) & MPJPE ($\downarrow$) \\
\midrule
D-Grasp & 58.17 & 0.599 & 36.18 \\
GraspXL (\textbf{Ours}) & \textbf{49.98} & \textbf{0.645} & \textbf{30.96} \\
\bottomrule
\end{tabular}
\label{tab:stageI}
\end{table}

\begin{table}[hbp] 
\centering
\caption{Ablation on Hand Representation}
\label{tab:hand_rep}
\resizebox{\linewidth}{!}{
\begin{tabular}{lccc} 
\toprule
Hand Representation & FVD ($\downarrow$) & PSNR ($\uparrow$) & MPJPE ($\downarrow$) \\
\midrule
Mesh Projection & 29.33 & \textbf{30.17} & 36.18 \\
Keypoints (\textbf{Ours}) & \textbf{29.13} & 30.05 & \textbf{30.96} \\
\bottomrule
\end{tabular}
} 
\end{table}

\label{sec:supp_setting}
\textbf{Training Details}: Our model was trained on a setup consisting of 8 x NVIDIA 800 GPUs, with a batch size of 4 x 8 and a learning rate of $1 \times 10^{-4}$. The training process involved 8,000 training steps, using the AdamW optimizer and the DeepSpeed training architecture~\citep{rajbhandari2020zero}. 

\textbf{Evaluation Details}: For the evaluation of video generation, we sample 1,600 videos, each consisting of 49 frames, from the test set. For the evaluation of Mean Per Joint Position Error (MPJPE), we utilize Hamer~\citep{pavlakos2024reconstructing} to estimate the hand joints in the generated videos, and compute the loss by comparing the estimated joint positions with the ground truth hand joints. To assess the performance on downstream tasks, we train the SimpleHand model for 200 epochs using its official implementation.

\textbf{Downstream Validation Details:} For the downstream task, given the input, we first leverage the appearance generator (the controllable image diffusion model) to randomly sample 30 candidates. Subsequently, we filter out samples with low-accuracy hand poses using Hamer. Specifically, we predict the hand keypoints using Hamer for every frame, compare them with the ground truth, and discard the bottom 25\% of the generated videos based on pose accuracy.

\section{Additional Results}

\subsection{Compute and Memory Benchmarking}

As shown in Table~\ref{sup:resources}, we report per-stage resource usage on an NVIDIA H20 GPU. Stage~I is lightweight, Stage~II has the highest peak memory (41.4~GB), and Stage~III is the slowest (245.7~s) due to diffusion computation. Overall, the full pipeline runs in \textbf{301.1~s} for 40 frames.

\subsection{Ablation on Masking Probability and Input Condition Quality}

We evaluate the robustness of our model by introducing random Gaussian noise into the input conditions. This noise is added to simulate real-world perturbations and assess how well the model can maintain performance despite such distortions. The results, as shown in Table~\ref{tab:mask}, highlight a noticeable performance degradation when no masking is applied — the model's performance metrics decrease significantly when exposed to noise.

However, when masking is applied to the input conditions, the performance drops are considerably reduced. This observation strongly suggests that the random masking technique enhances the model's robustness. By masking portions of the input, the model appears to be less sensitive to noise, likely because it focuses on more stable, less noisy features of the data. Therefore, the use of random masking improves the model's generalization ability and resilience to external noise, thus supporting the hypothesis that masking is an effective strategy for improving robustness in challenging conditions.

\subsection{Ablation on Appearance Generation Method}

To evaluate the sensitivity of our method to Stage-I trajectories, we substituted GraspXL with D-Grasp. As shown in Table~\ref{tab:stageI}, GraspXL achieves superior performance, demonstrating that the final generation quality relies on the quality of Stage-I pose sequence.

\subsection{Ablation on Hand Representations}
We ablate hand keypoints versus 2D hand mesh projections as conditioning. Table~\ref{tab:hand_rep} shows that keypoints perform better on most metrics, especially on MPJPE. This is possibly because mesh projections are more prone to self-occlusion.

\subsection{Ablation on Hand Encoder}

\begin{table}[hbp]
\centering
\caption{Ablation on Hand Encoder}
\begin{tabular}{c|ccc}
\toprule
Encoder Type & FVD ($\downarrow$) & PSNR ($\uparrow$) & MPJPE ($\downarrow$) \\
\midrule
MLP & 31.59 & 30.07 & 21.96 \\
VAE (\textbf{Ours}) & \textbf{29.13} & \textbf{30.17} & \textbf{19.37} \\
\bottomrule
\end{tabular}
\label{tab:hand_enc}
\end{table}

To validate the VAE-based condition encoder, we evaluate it on 1,000 sampled keypoint images and obtain a PSNR of 40.58, indicating a low reconstruction error for the conditioning images. As shown in Table~\ref{tab:hand_enc}, the VAE outperforms an MLP-based 2D coordinate encoder across all metrics. This superior performance is primarily attributed to the VAE's enhanced ability to preserve local spatial information.

\subsection{Ablation on I2V Backbone}

\begin{table}[hbp] 
\centering
\caption{Ablation on Backbone}
\label{tab:abl_backbone}
\resizebox{\linewidth}{!}{
\begin{tabular}{lccc} 
\toprule
Encoder Type & FVD ($\downarrow$) & PSNR ($\uparrow$) & MPJPE ($\downarrow$) \\
\midrule
InterDyn (SVD w/ single cond) & 38.83 & 24.86 & 28.15 \\
SVD w/ multi conds & \underline{34.91} & \underline{25.84} & \underline{25.11} \\
\textbf{Ours} (CogVideo w/ multi conds) & \textbf{29.13} & \textbf{30.17} & \textbf{19.37} \\
\bottomrule
\end{tabular}
} 
\end{table}

\begin{figure*}
\centering
\includegraphics[width=1\textwidth]{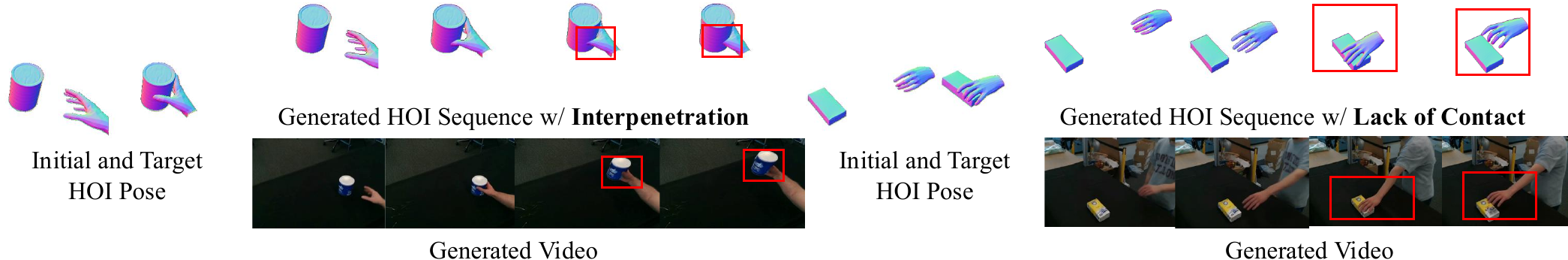}
\caption{More qualitative results on DexYCB dataset (a).
}
\label{fig:inves}
\end{figure*}

To isolate the impact of our trimodal conditioning, we present a controlled ablation in Table~\ref{tab:abl_backbone}. Notably, even when employing a comparatively weaker backbone, the multi-conditioned SVD~\cite{blattmann2023stable} baseline consistently outperforms the single-condition InterDyn across all metrics. This comparison demonstrates that integrating multiple conditions significantly enhances the overall generation performance.

\subsection{Investigation in Error Propagation}

Experiments in Figure~\ref{fig:inves} show that Stage-I geometric errors (e.g., interpenetration or missing contact) can propagate, leading to physically implausible interactions even if the generated video appears photorealistic. Furthermore, we observe that Stage-III quality heavily relies on the appearance guidance from Stage II: low-quality initial reference frames degrade the final textures and exacerbate temporal flickering.

\subsection{Additional Qualitative Results}
\begin{figure*}
\centering
\includegraphics[width=1\textwidth]{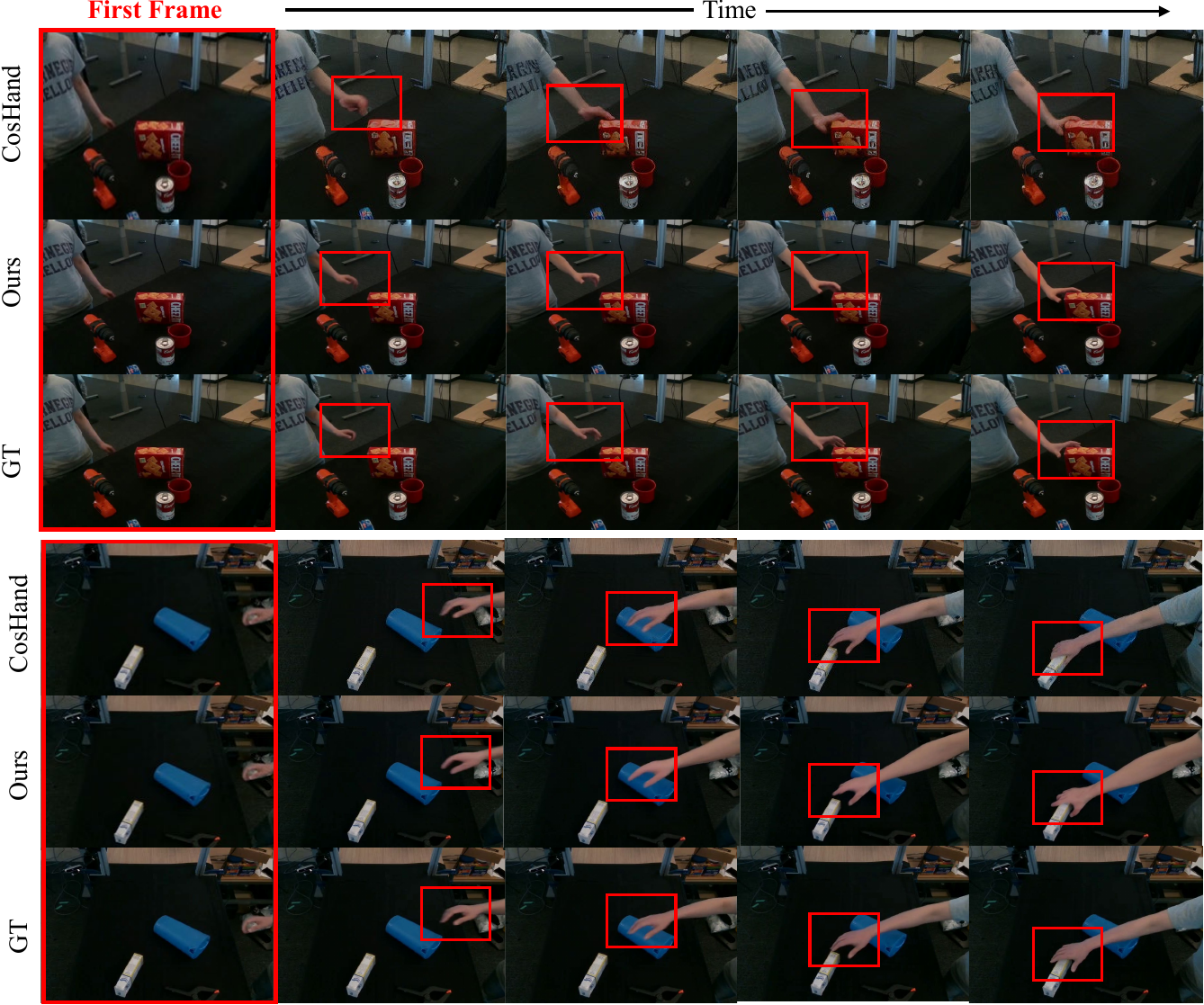}
\caption{More qualitative results on DexYCB dataset (a).
}
\label{fig:supp_1}
\end{figure*}

\begin{figure*}
\centering
\includegraphics[width=1\textwidth]{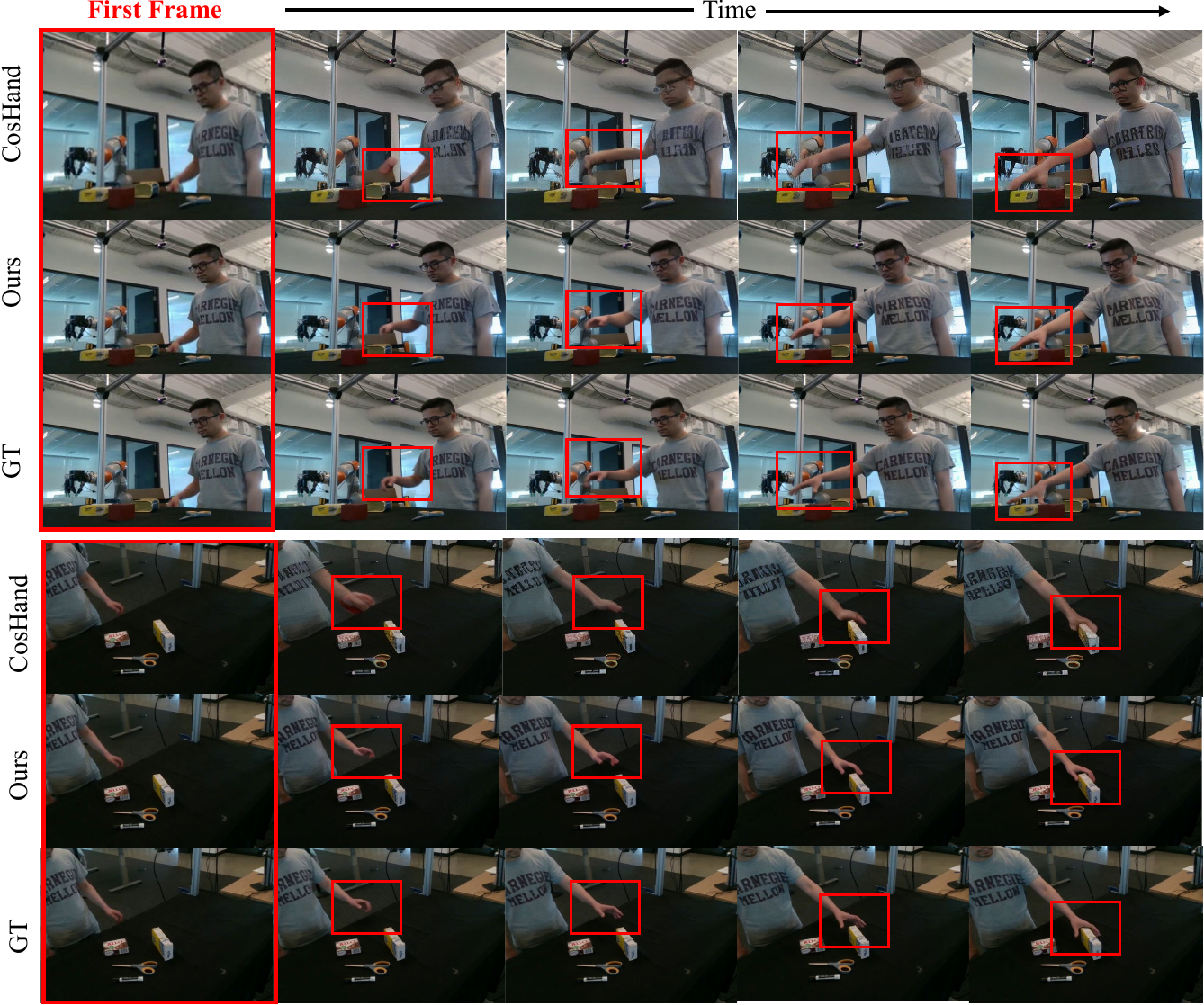}
\caption{More qualitative results on DexYCB dataset (b).
}
\label{fig:supp_2}
\end{figure*}

\begin{figure*}
\centering
\includegraphics[width=1\textwidth]{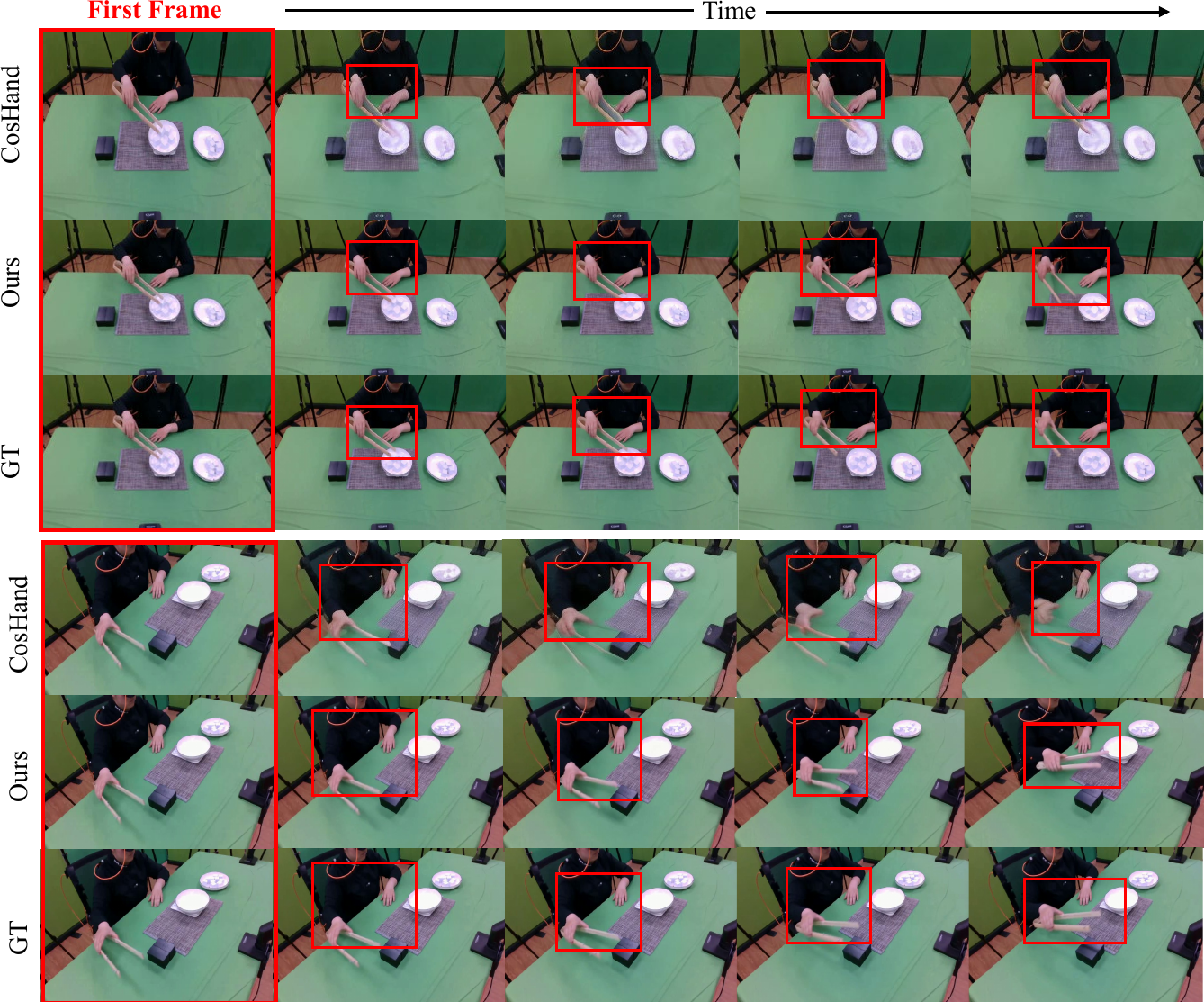}
\caption{More qualitative results on OAKINK2 dataset (a).
}
\label{fig:supp_3}
\end{figure*}

\begin{figure*}
\centering
\includegraphics[width=1\textwidth]{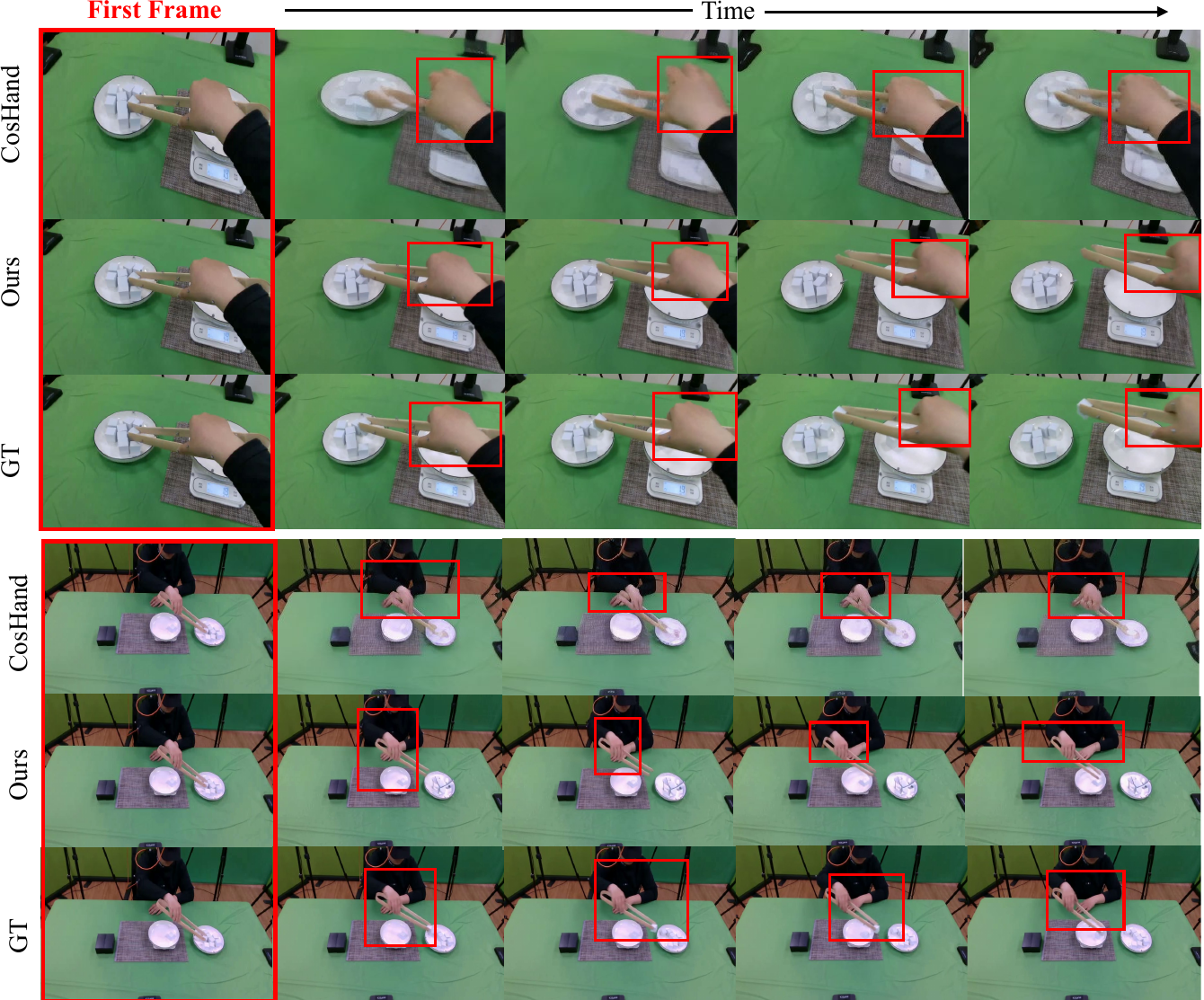}
\caption{More qualitative results on OAKINK2 dataset (b).
}
\label{fig:supp_4}
\end{figure*}

We provide more qualitative results in Figure~\ref{fig:supp_1} and Figure~\ref{fig:supp_2} for DexYCB dataset, Figure~\ref{fig:supp_3} and Figure~\ref{fig:supp_4} for OANINK2 dataset.

\end{document}